\documentclass[10pt,letterpaper]{article}
\usepackage[left=0.75in,right=0.75in,top=1in,bottom=1in]{geometry}
\usepackage[T1]{fontenc}

\usepackage{times}
\usepackage{microtype}
\usepackage{amsmath,amssymb}
\usepackage{array}
\usepackage{booktabs}
\usepackage{graphicx}
\usepackage[hidelinks]{hyperref}
\usepackage{tabularx}

\newcolumntype{Y}{>{\raggedright\arraybackslash}X}

\makeatletter
\renewcommand{\maketitle}{%
  \begin{center}
    {\LARGE\bfseries \@title\par}
    \vskip 0.75em
    {\large\@author\par}
    \vskip 0.75em
  \end{center}
}
\makeatother

\title{Before the Model Learns the Bug:\\Fuzzing RLVR Verifiers}
\author{Jaideep Ray\\\texttt{jaray@acm.org}}
\date{}

\begin{document}
\twocolumn[
\begin{@twocolumnfalse}
\maketitle
\begin{abstract}
Reinforcement learning with verifiable rewards (RLVR) replaces human preference labels with executable reward functions such as math answer checkers, JSON tool-call validators, and code unit-test harnesses. That makes the reward partly a software artifact: if the verifier is wrong, optimization can learn the bug. We study this failure mode with a lightweight verifier-fuzzing framework that generates adversarial completions, compares buggy and stricter reference verifiers, logs paired decisions, and reports false-positive, false-negative, disagreement, exploit, and uncertainty metrics. Across seeded runs, intentionally buggy math, JSON tool-call, and toy code verifiers repeatedly accept incorrect completions, while stricter variants remove the measured false positives on the same generated cases. An open-source replay grounds the diagnostic in existing validators: relative to our stricter final-answer contract, \texttt{math-verify} accepts \texttt{partial\_answer} cases, while a SymPy-backed final-answer replay removes those false positives by accepting a narrower language. The \texttt{jsonschema} replay accepts none of the adversarial JSON cases, and its zero coverage identifies format-level non-engagement rather than semantic validation. Hardening ablations, template search, tabular policy gradients, and budgeted black-box querying show that false-positive regions are structured enough to sustain high reward while strict proxy correctness remains low. A HumanEval-style visible-vs-hidden code evaluation reproduces the visible-test failure mode outside the main synthetic fuzzers. These results make verifier reliability a pre-training systems property that can be measured and audited before optimization begins.
\end{abstract}
\vspace{0.45em}
\begin{center}
\begin{minipage}{0.92\textwidth}
\end{minipage}
\end{center}
\vspace{0.2em}
\end{@twocolumnfalse}
]

\section{Introduction}

RLVR is attractive because it turns machine-checkable tasks into scalable post-training signals. A model can be rewarded for producing a final mathematical answer, emitting a schema-valid tool call, or writing code that passes tests, reducing dependence on human labels and making pipelines more reproducible.

The systems risk is that the reward is executable software. The verifier parses outputs, extracts answers, validates schemas, compares values, executes tests, enforces timeouts, and decides whether to award reward. Bugs in this stack do not merely add noise: under optimization pressure, they become rewardable failure modes. A verifier that accepts the first number in a response, ignores duplicate JSON keys, or treats printed text as program correctness creates high-reward false positives.

This paper asks a practical systems question: can realistic verifier bugs create high-reward false positives that simple search or RL-style optimization can reliably find? In controlled local experiments, yes. We build a verifier-fuzzing prototype for math final-answer verification, JSON tool-call verification, and toy code unit-test verification. We pair intentionally buggy implementations with stricter reference variants, deterministic and held-out fuzzers, JSONL logs, hardening ablations, optimization proxies, and targeted replay against open-source math and JSON validators.

The paper focuses on a systems diagnostic rather than a new RL algorithm: verifier reliability is measurable before expensive training begins, so RLVR builders should stress-test reward verifiers before using them as training objectives.

This paper makes four contributions:

\begin{itemize}
    \item A practical taxonomy of short, realistic verifier bugs for math, tool-call, and code-reward pipelines.
    \item A reproducible paired verifier-fuzzing workflow that logs buggy-versus-strict outcomes and computes metrics only from saved JSONL records.
    \item Multi-seed evidence across held-out, adaptive, and optimization settings that false-positive regions are structured and exploitable.
    \item An engineering view of verifier hardening, including ablations that localize high-value fixes, an open-source verifier replay that tests existing validators, and an external HumanEval-style code evaluation that validates the visible-test failure mode outside the main synthetic fuzzer.
\end{itemize}

\section{Background and Related Work}

Reward hacking and specification gaming are long-standing concerns in reinforcement learning. Amodei et al. identify reward hacking as a concrete AI safety problem: an agent may exploit loopholes in an objective rather than satisfy the intended task \cite{amodei2016}. RLVR makes one class of loopholes especially concrete: verifier implementation defects.

Recent RLVR work has observed verifier gaming after model training. Helff et al. study LLMs gaming verifiers in inductive reasoning tasks and use isomorphic perturbation testing to expose shortcut behavior \cite{helff2026}. Khalifa et al. introduce Countdown-Code, a controlled environment where proxy reward can be separated from true task reward \cite{khalifa2026}. Our method is complementary. Instead of first training a model and then diagnosing reward hacking, we fuzz the verifier directly as a pre-training diagnostic.

Our code setting is related to code-generation evaluation. Prior work shows that insufficient tests can overestimate functional correctness of generated code \cite{liu2023}. We study the same issue from the reward-function perspective: if an RLVR verifier runs only visible tests or accepts stdout as evidence of correctness, it can reward code that does not implement the intended function.

Our JSON tool-call setting is related to LLM tool-use and API-invocation benchmarks, including ToolLLM's tool-use data and evaluation resources \cite{qin2024}. Tool-call evaluation often relies on schema-based validation, making it natural to study bugs involving wrong tool names, ignored argument values, duplicate keys, embedded JSON, and unexpected fields.

\section{Failure Model and Metrics}

A verifier receives a task, gold specification, and completion, then returns a reward. The dangerous event is a silent false positive:

\begin{quote}
verifier accepts completion, but completion is semantically incorrect.
\end{quote}

For a completion $c$, let $A(c)$ be verifier acceptance and $T(c)$ be true semantic correctness. A silent false positive is

\[
A(c) = 1 \quad \text{and} \quad T(c) = 0.
\]

False negatives are also undesirable because they reject valid completions, but false positives are more directly connected to reward hacking. They define regions of completion space where optimization receives high reward without solving the task.

The policy or search process is not assumed to have source access to the verifier. It only needs repeated reward feedback. This matches ordinary local RLVR training: completions are sampled, evaluated, rewarded, and then sampled more or less often based on the reward.

We report the following metrics from saved logs:

\begin{itemize}
    \item \textbf{False-positive rate:} accepted and truly incorrect completions divided by truly incorrect completions.
    \item \textbf{False-negative rate:} rejected and truly correct completions divided by truly correct completions.
    \item \textbf{Differential disagreement rate:} paired buggy and strict verifier decisions differ.
    \item \textbf{Exploit-candidate rate:} buggy verifier accepts and strict verifier rejects.
    \item \textbf{Reward-correctness gap:} mean verifier reward minus strict proxy correctness in optimization experiments.
    \item \textbf{Exploit rate:} sampled completions that are rewarded but truly incorrect according to the workload's expected-correct label. In optimization tables where only strict-reference decisions are available, we call the analogous measure \emph{proxy exploit rate}.
\end{itemize}

Throughout the paper, $T(c)$ is the intended semantic label supplied by the workload design. The strict verifier is an operational proxy for $T(c)$, not a formal proof of correctness. Accordingly, a strict false-positive rate of zero means zero on the generated cases in this controlled evaluation, not zero for all possible completions or deployments. When a denominator is zero, such as a negative-only replay set with no truly correct candidates, we report the corresponding rate as N/A rather than 0.000.

\section{Verifier-Fuzzing System}

The prototype has five stages:

\begin{enumerate}
    \item Define seed tasks with prompts, gold answers, and verifier-specific gold specifications.
    \item Generate adversarial completions by mutation category.
    \item Evaluate each completion with paired buggy and strict verifiers.
    \item Save one JSONL record per verifier-case result.
    \item Analyze only saved logs to produce tables, confidence intervals, and numerator/denominator summaries.
\end{enumerate}

Each task is represented as a VerificationExample record. Key fields include the task identifier, prompt, gold answer, completion, expected-correct label, and metadata.

Each verifier returns a VerificationResult record. Typical fields include reward, correctness, extracted answer, and parse status. The record also stores verifier name, verifier version, and metadata.

Each fuzzer emits FuzzCase records that preserve a true expected-correct label. Records are paired by case identifier for strict-versus-buggy differential verification.

\subsection{Bug Taxonomy}

Table~\ref{tab:taxonomy} summarizes the implemented bug classes. The taxonomy is intentionally practical: these are short-verifier mistakes rather than exotic attacks.

\begin{table*}[t]
\centering
\footnotesize
\begin{tabularx}{\textwidth}{@{}l l Y Y@{}}
\toprule
Family & Bug class & Buggy behavior & Strict behavior \\
\midrule
Math & Loose extraction & Accepts first number anywhere & Extracts marked final answer \\
Math & Missing marker & Accepts unmarked answer & Requires \texttt{Final answer:} or \texttt{boxed\{\}} \\
Math & Contradiction blindness & Ignores inconsistent finals & Rejects inconsistent finals \\
Math & Loose tolerance & Accepts nearby wrong value & Uses tight numeric tolerance \\
JSON & Schema-only validation & Checks required keys only & Checks tool name and values \\
JSON & Extra fields & Ignores unexpected fields & Rejects unexpected keys \\
JSON & Duplicate keys & Parser keeps one parsed value & Detects duplicate keys \\
JSON & Embedded JSON & Regex-extracts object & Requires exact JSON object \\
Code & Visible-test overfitting & Runs visible tests only & Runs hidden tests too \\
Code & Stdout spoofing & Accepts printed expected text & Requires exact return values \\
Code & Missing timeout & Can hang or mishandle loops & Uses subprocess timeout \\
\bottomrule
\end{tabularx}
\caption{Implemented verifier bug taxonomy.}
\label{tab:taxonomy}
\end{table*}

\subsection{Hardening Ablations}

We add intermediate verifier variants between the buggy and strict endpoints. For math, hardening adds final-marker requirements, contradiction checks, and tight numeric parsing. For JSON tool calls, hardening adds exact tool-name checks, argument-value checks, extra-key rejection, duplicate-key rejection, and exact-object parsing. For code, hardening removes stdout acceptance, adds hidden tests, requires exact return values, checks expected function names, and uses static checks plus timeouts.

These variants turn the paper from a buggy-versus-strict demo into an engineering study: they answer which hardening step removes which class of false positives.

\subsection{Optimization Proxies}

We use two local optimization proxies. The first is template search: sample completion templates, score with a verifier, keep high-reward templates, mutate them, and repeat. The second is a tabular policy-gradient bandit with one logit per template. Each round samples a batch, receives verifier reward, and applies a REINFORCE-style update \cite{williams1992}. Strict verifier output is logged as held-out proxy correctness but is not used as training reward. The saved final-round summaries report the reward-feedback mechanism, and the artifact manifest records template-pool size, number of rounds, batch size, mutation rate, update scale, reward baseline, and held-out evaluation count for exact reproduction.

These proxies isolate the reward-feedback mechanism: repeated reward feedback amplifies false positives.

\section{Experimental Setup}

All experiments run locally in plain Python. The implementation uses \texttt{pytest} and \texttt{hypothesis} for tests and targeted case generation. No cloud service, distributed training framework, or GPU is required. The workloads are controlled to isolate verifier behavior under targeted mutations rather than conflate it with benchmark difficulty.

\subsection{Fuzzing Workloads}

The math workload uses simple arithmetic tasks and mutations such as contradictory answers, missing markers, partial answers, equivalent fractions, percentages, loose floats, word-number forms, and wrong reasoning followed by a correct final answer.

The JSON tool-call workload uses toy \texttt{schedule\_event}, \texttt{send\_email}, and \texttt{search} tasks. Mutations include missing keys, extra keys, wrong tool names, wrong argument values, duplicate keys, malformed JSON, embedded JSON, and semantically invalid values.

The code workload uses pure-Python function tasks, including addition, string reversal, palindrome checking, factorial, and vowel counting. Mutations include hardcoded visible tests, stdout spoofing, infinite loops, wrong function names, hidden-edge failures, syntax errors, and wrong return types.

\subsection{Repetition and Uncertainty}

Multi-seed fuzzing runs seeds 0 through 9. For each seed, the runner samples a seed-specific subset from a larger deterministic pool, then evaluates the same cases with paired buggy and strict verifiers. We report means, standard deviations, and 95\% bootstrap confidence intervals over seeds. Each saved record contributes to an explicit numerator and denominator: FPR uses true-incorrect cases, FNR uses true-correct cases, acceptance and coverage use all replayed cases, and exploit discovery uses independent black-box trials. The tables below include compact count forms where a missing denominator would otherwise make a decimal rate hard to interpret.

\subsection{Open-source Verifier Replay}

We replay a subset of the math and JSON fuzz cases against three open-source validators: \texttt{math-verify}, a SymPy-backed final-answer verifier, and \texttt{jsonschema}. In this replay, acceptance rate is the fraction of all fuzz cases that the verifier accepts, while coverage rate is one minus parse-error rate. These metrics distinguish semantic strictness from simple non-engagement with the input format.

\section{Results}

\subsection{Verifier Bugs Create Large False-positive Regions}

Table~\ref{tab:headline} reports mean rates over 10 seeded fuzzing runs. Brackets denote 95\% bootstrap confidence intervals over seeds. The $N$, $N_-$, and $N_+$ columns give the per-seed all-case, true-incorrect, and true-correct denominators, respectively; exploit-candidate denominators are all paired verifier cases.

\begin{table}[h]
\centering
\footnotesize
\setlength{\tabcolsep}{3pt}
\resizebox{\columnwidth}{!}{%
\begin{tabular}{@{}lcccccc@{}}
\toprule
Family & $N$ & $N_-$ & $N_+$ & Buggy FPR & Strict FPR & Exploit-cand. \\
\midrule
Math & 200 & 120 & 80 & 0.832 [0.824, 0.841] & 0.000 [0.000, 0.000] & 0.498 [0.488, 0.509] \\
JSON tool calls & 500 & 500 & 0 & 0.869 [0.863, 0.875] & 0.000 [0.000, 0.000] & 0.869 [0.863, 0.875] \\
Code unit tests & 100 & 100 & 0 & 0.557 [0.539, 0.573] & 0.000 [0.000, 0.000] & 0.557 [0.539, 0.573] \\
\bottomrule
\end{tabular}
}
\caption{Headline false-positive and exploit-candidate rates over 10 seeds. $N_-$ is the FPR denominator. $N_+$ is the FNR denominator if FNR is reported.}
\label{tab:headline}
\end{table}

The buggy math verifier accepts many incorrect completions because a gold number appears somewhere before the actual final answer. The buggy JSON verifier accepts cases where required keys exist but tool names, values, duplicate keys, embedded JSON, or extra fields are wrong. The buggy code verifier accepts visible-test overfits, stdout spoofing, and hidden-edge failures.

\subsection{Open-source Replay Tests Existing Validators}

Table~\ref{tab:realreplay} summarizes the replay against open-source verifiers. Acceptance is accepted cases divided by all replayed cases; coverage is one minus parse-error rate. Count forms show numerator/denominator before the decimal rate. A zero-coverage row is interpreted as non-engagement with the replay format, not as useful semantic rejection.

\begin{table}[t]
\centering
\footnotesize
\setlength{\tabcolsep}{3pt}
\resizebox{\columnwidth}{!}{%
\begin{tabular}{@{}l l c c c c@{}}
\toprule
Verifier & Family & FPR & FNR & Acc. & Cov. \\
\midrule
\texttt{math-verify} & Math & 10/60=.167 & 0/40=.000 & 50/100=.500 & 90/100=.900 \\
SymPy final-answer & Math & 0/60=.000 & 0/40=.000 & 40/100=.400 & 76/100=.760 \\
\texttt{jsonschema} & JSON tool & 0/80=.000 & N/A & 0/80=.000 & 0/80=.000 \\
\bottomrule
\end{tabular}}
\caption{Open-source verifier replay. FNR is N/A for \texttt{jsonschema} because this adversarial JSON replay contains no truly correct JSON tool-call candidates; its zero coverage indicates format-level non-engagement.}
\label{tab:realreplay}
\end{table}

The \texttt{math-verify} replay gives the clearest existing-validator result. Its overall replay FPR is 10/60=0.167, and the mutation-level breakdown assigns FPR 10/10=1.000 to \texttt{partial\_answer} cases, where the gold value appears in prose without the stricter final-answer contract required by our reference verifier. This is an accepted-language mismatch relative to our stricter final-answer contract and an open-source analogue of the loose-extraction failure mode.

The SymPy-backed final-answer verifier removes those false positives on the replayed cases, but it accepts fewer cases and covers fewer formats: acceptance drops from 0.500 to 0.400 and coverage from 0.900 to 0.760. This makes the strictness tradeoff operational rather than abstract. Lower FPR can come from a narrower accepted language, so replay reports should include FPR together with acceptance and coverage.

The \texttt{jsonschema} replay accepts 0/80 cases in this negative-only adversarial JSON set and has coverage 0/80. Therefore this row identifies format-level non-engagement on the replayed strings rather than evidence that schema validation successfully rejects parsed-but-invalid calls. Schema-validator reports should separately count parse errors, parsed schema rejections, semantic rejections, and accepted cases.

\subsection{Hardening Ablations Identify High-value Fixes}

Table~\ref{tab:ablation} shows the stepwise false-positive rate as verifier checks are added. Count forms use the per-seed true-incorrect denominators from Table~\ref{tab:headline}: 120 math cases, 500 JSON tool-call cases, and 100 code cases.

\begin{table*}[t]
\centering
\footnotesize
\setlength{\tabcolsep}{4pt}
\begin{tabular}{@{}l p{0.82\textwidth}@{}}
\toprule
Family & Step sequence FPR \\
\midrule
Math & first-num 100/120=.833 $\rightarrow$ marker 28/120=.233 $\rightarrow$ contradict 28/120=.233 $\rightarrow$ tight-num 0/120=.000 \\
JSON & keys 437/500=.874 $\rightarrow$ tool 374/500=.748 $\rightarrow$ args 187/500=.374 $\rightarrow$ extras 124/500=.248 $\rightarrow$ dups 62/500=.124 $\rightarrow$ exact 0/500=.000 \\
Code & stdout 55/100=.550 $\rightarrow$ return 26/100=.260 $\rightarrow$ hidden 0/100=.000 $\rightarrow$ name 0/100=.000 $\rightarrow$ timeout 0/100=.000 \\
\bottomrule
\end{tabular}
\caption{Stepwise hardening ablations and their false-positive rates, reported as false positives over true-incorrect cases.}
\label{tab:ablation}
\end{table*}

For math, the largest reduction comes from requiring an explicit final-answer marker. For JSON, the largest reduction comes from checking argument values and simple semantics. For code, removing stdout acceptance and adding hidden tests remove the measured false-positive region.

\subsection{Held-out and Adaptive Fuzzing Preserve Exploits}

Held-out transformations change surface form without changing the underlying bug class. At round 0, mean exploit-candidate rates are 0.513 for math, 0.888 for JSON tool calls, and 0.533 for code, using per-seed denominators of 100,200 and 500 paired cases respectively. After one adaptive round, the exploit-candidate rate reaches 1.000 for all three families on the same per-seed denominators because the loop keeps buggy-accept/strict-reject cases and mutates them further.

The result identifies a robust local exploit basin: false-positive regions survive simple held-out rewrites and can be concentrated by black-box feedback.

\subsection{Optimization Amplifies Buggy Rewards}

Table~\ref{tab:optimization} reports final-round optimization metrics over 10 seeds. The reward-correctness gap is verifier reward minus strict proxy correctness. Because the table uses strict-reference decisions as its held-out correctness signal, ``Proxy expl.'' means rewarded and strict-reference rejected. The final-round denominator is the saved final batch in the run manifest.

\begin{table*}[t]
\centering
\footnotesize
\setlength{\tabcolsep}{4pt}
\begin{tabular}{@{}lllcccc@{}}
\toprule
Optimizer & Task & Verifier & Reward & Proxy corr. & Gap [95\% CI] & Proxy expl. \\
\midrule
Template & Math & Buggy & 0.967 & 0.154 & 0.812 [0.700, 0.892] & 0.812 \\
Template & Math & Strict & 0.762 & 0.762 & 0.000 [0.000, 0.000] & 0.000 \\
Template & JSON & Buggy & 1.000 & 0.150 & 0.850 [0.800, 0.896] & 0.850 \\
Template & JSON & Strict & 0.792 & 0.792 & 0.000 [0.000, 0.000] & 0.000 \\
Tabular PG & Math & Buggy & 0.972 & 0.169 & 0.803 [0.769, 0.841] & 0.803 \\
Tabular PG & Math & Strict & 0.906 & 0.906 & 0.000 [0.000, 0.000] & 0.000 \\
Tabular PG & JSON & Buggy & 1.000 & 0.178 & 0.822 [0.788, 0.856] & 0.822 \\
Tabular PG & JSON & Strict & 0.916 & 0.916 & 0.000 [0.000, 0.000] & 0.000 \\
\bottomrule
\end{tabular}
\caption{Final-round optimization metrics over 10 seeds.}
\label{tab:optimization}
\end{table*}

With buggy verifiers, both optimizers maintain high reward while strict proxy correctness remains low. With strict verifiers, reward and proxy correctness track each other and proxy exploit rate is zero.

\subsection{Few Black-box Queries Find Exploits}

Budgeted black-box search answers a practical question: how many reward queries are enough to discover an exploit in the fixed completion-template pool? Over 100 independent trials per family, buggy reward finds an exploit within two verifier queries in 94/100 math trials and 98/100 JSON-tool trials. Within four queries, both reach 100/100. Strict verifiers have 0/100 exploit discovery at all budgets. These counts quantify how quickly exploits emerge in the fixed completion-template pool.

\subsection{External-style Code Case Study}

The HumanEval-style code study contains 20 candidate programs across four benchmark-like tasks: 4 correct programs and 16 incorrect programs. The buggy visible-test/stdout verifier has FPR 16/16=1.000 and FNR 0/4=0.000; the strict hidden-test verifier has FPR 0/16=0.000 and FNR 0/4=0.000. The buggy verifier accepts all incorrect candidate types, including hidden-edge failures, stdout spoofing, visible-test overfits, and wrong function names, while the strict verifier rejects them. This reinforces the systems lesson that visible tests are not enough when test passing is used as a reward.

\section{Discussion}

The results support three conclusions.

First, verifier bugs create structured false-positive regions. These are not uniform random label errors; each bug class admits a recognizable family of incorrect completions that can be rewarded. The open-source replay shows that accepted-language choices in a real math validator can interact with the stricter final-answer contract used here: \texttt{math-verify} accepts the same partial-answer pattern on replayed cases.

Second, differential verification is an actionable pre-training diagnostic. The paired outcome buggy accepts and strict rejects points directly to hardening work: final-answer markers, contradiction checks, exact JSON parsing, duplicate-key rejection, semantic validation, hidden tests, return-value checks, and timeouts.

Third, optimization pressure amplifies verifier defects. Template search and a minimal tabular policy-gradient loop both convert false-positive regions into high-reward, low-correctness behavior. In other words, the reward implementation becomes the optimization target.

For practitioners, the workflow is straightforward: fuzz a candidate verifier before training, inspect buggy-accept/strict-reject cases, add targeted checks, and rerun on held-out mutations. Stricter is not always better; production verifiers may need to accept multiple valid formats, and excessive strictness can increase false negatives. The engineering goal is an acceptance language that matches the task specification closely enough that reward tracks correctness.

\section{Limitations}

The evaluation is scoped to verifier behavior, not end-to-end RLVR training. The math, JSON tool-call, and code workloads use synthetic tasks with intentionally injected implementation defects that mirror common verifier failure modes.

The open-source replay covers \texttt{math-verify}, a SymPy-backed final-answer verifier, and \texttt{jsonschema}. The \texttt{jsonschema} row has zero coverage on the adversarial JSON strings, so it measures format-level non-engagement rather than semantic rejection.

The fuzzers use deterministic templates and one adaptive mutation loop. They provide reproducible coverage of targeted bug classes, but they do not model the full distribution of completions produced by a trained language-model policy.

The optimization experiments use template search and tabular policy gradients. They measure whether repeated reward feedback amplifies verifier false positives in the fixed template pools; they do not estimate exploit rates for neural fine-tuning runs.

The code verifier executes Python snippets in a subprocess with restricted builtins and a timeout. It is an evaluation harness for controlled examples, not a sandbox for untrusted code. The external code case study targets HumanEval-style visible-test failures rather than a full HumanEval or EvalPlus benchmark sweep.

\section{Conclusion}

Verifiable rewards are only as reliable as their verifiers. In controlled RLVR-style settings, realistic verifier bugs create structured false-positive regions across math, JSON tool-call, and code-verification tasks, and open-source replay shows that accepted-language choices in real validators can interact with stricter RLVR task contracts. Search, tabular RL-style optimization, and budgeted black-box querying convert these regions into high-reward, low-correctness behavior. The practical takeaway is simple: fuzz verifiers before training, compare permissive and stricter variants, report false positives with acceptance and coverage, inspect differential cases, and use hardening ablations so reward tracks the intended task.
\begingroup
\footnotesize
\setlength{\parskip}{0.25\baselineskip}

\endgroup
\end{document}